\documentclass{article} 
\usepackage{iclr2018_conference,times}
\usepackage{hyperref}
\usepackage{url}
\usepackage{graphicx}
\usepackage{pgfplots}
\usepackage{caption,subcaption}
\pgfplotsset{compat=1.5}

\usepackage{amsmath}
\usepackage{amssymb}
\usepackage{bm}
\usepackage{multirow}
\usepackage{xspace}

\newcommand{\safemath}[2]{\newcommand{#1}{\ensuremath{#2}\xspace}}

\newcommand{\e}{\mathbf{e}}
\newcommand{\f}{\mathbf{f}}

\renewcommand{\u}{\bm{u}}
\newcommand{\x}{\bm{x}}
\renewcommand{\f}{\bm{f}}
\newcommand{\y}{\bm{y}}
\newcommand{\z}{\bm{z}}

\newcommand{\R}{\mathbb{R}}

\safemath{\Ab}{\mathbb{A}}
\safemath{\Bb}{\mathbb{B}}
\safemath{\Cb}{\mathbb{C}}
\safemath{\Db}{\mathbb{D}}
\safemath{\Eb}{\mathbb{E}}
\safemath{\Fb}{\mathbb{F}}
\safemath{\Gb}{\mathbb{G}}
\safemath{\Hb}{\mathbb{H}}
\safemath{\Ib}{\mathbb{I}}
\safemath{\Jb}{\mathbb{J}}
\safemath{\Kb}{\mathbb{K}}
\safemath{\Lb}{\mathbb{L}}
\safemath{\Mb}{\mathbb{M}}
\safemath{\Nb}{\mathbb{N}}
\safemath{\Ob}{\mathbb{O}}
\safemath{\Pb}{\mathbb{P}}
\safemath{\Qb}{\mathbb{Q}}
\safemath{\Rb}{\mathbb{R}}
\safemath{\Sb}{\mathbb{S}}
\safemath{\Tb}{\mathbb{T}}
\safemath{\Ub}{\mathbb{U}}
\safemath{\Vb}{\mathbb{V}}
\safemath{\Wb}{\mathbb{W}}
\safemath{\Xb}{\mathbb{X}}
\safemath{\Yb}{\mathbb{Y}}
\safemath{\Zb}{\mathbb{Z}}

\safemath{\Af}{\mathbf{A}}
\safemath{\Bf}{\mathbf{B}}
\safemath{\Df}{\mathbf{D}}
\safemath{\Ef}{\mathbf{E}}
\safemath{\Ff}{\mathbf{F}}
\safemath{\Gf}{\mathbf{G}}
\safemath{\Hf}{\mathbf{H}}
\safemath{\Jf}{\mathbf{J}}
\safemath{\Kf}{\mathbf{K}}
\safemath{\Lf}{\mathbf{L}}
\safemath{\Mf}{\mathbf{M}}
\safemath{\Nf}{\mathbf{N}}
\safemath{\Of}{\mathbf{O}}
\safemath{\Pf}{\mathbf{P}}
\safemath{\Qf}{\mathbf{Q}}
\safemath{\Rf}{\mathbf{R}}
\safemath{\Sf}{\mathbf{S}}
\safemath{\Tf}{\mathbf{T}}
\safemath{\Uf}{\mathbf{U}}
\safemath{\Vf}{\mathbf{V}}
\safemath{\Wf}{\mathbf{W}}
\safemath{\Xf}{\mathbf{X}}
\safemath{\Yf}{\mathbf{Y}}
\safemath{\Zf}{\mathbf{Z}}

\safemath{\af}{\mathbf{a}}
\safemath{\df}{\mathbf{d}}
\safemath{\ef}{\mathbf{e}}
\safemath{\ff}{\mathbf{f}}
\safemath{\gf}{\mathbf{g}}
\safemath{\hf}{\mathbf{h}}
\safemath{\jf}{\mathbf{j}}
\safemath{\kf}{\mathbf{k}}
\safemath{\lf}{\mathbf{l}}
\safemath{\mf}{\mathbf{m}}
\safemath{\nf}{\mathbf{n}}
\safemath{\of}{\mathbf{o}}
\safemath{\pf}{\mathbf{p}}
\safemath{\qf}{\mathbf{q}}
\safemath{\rf}{\mathbf{r}}
\safemath{\tf}{\mathbf{t}}
\safemath{\uf}{\mathbf{u}}
\safemath{\vf}{\mathbf{v}}
\safemath{\wf}{\mathbf{w}}
\safemath{\xf}{\mathbf{x}}
\safemath{\yf}{\mathbf{y}}
\safemath{\zf}{\mathbf{z}}

\safemath{\Ac}{\mathcal{A}}
\safemath{\Bc}{\mathcal{B}}
\safemath{\Cc}{\mathcal{C}}
\safemath{\Dc}{\mathcal{D}}
\safemath{\Ec}{\mathcal{E}}
\safemath{\Fc}{\mathcal{F}}
\safemath{\Gc}{\mathcal{G}}
\safemath{\Hc}{\mathcal{H}}
\safemath{\Ic}{\mathcal{I}}
\safemath{\Jc}{\mathcal{J}}
\safemath{\Kc}{\mathcal{K}}
\safemath{\Lc}{\mathcal{L}}
\safemath{\Mc}{\mathcal{M}}
\safemath{\Nc}{\mathcal{N}}
\safemath{\Oc}{\mathcal{O}}
\safemath{\Pc}{\mathcal{P}}
\safemath{\Qc}{\mathcal{Q}}
\safemath{\Rc}{\mathcal{R}}
\safemath{\Sc}{\mathcal{S}}
\safemath{\Tc}{\mathcal{T}}
\safemath{\Uc}{\mathcal{U}}
\safemath{\Vc}{\mathcal{V}}
\safemath{\Wc}{\mathcal{W}}
\safemath{\Xc}{\mathcal{X}}
\safemath{\Yc}{\mathcal{Y}}
\safemath{\Zc}{\mathcal{Z}}

\safemath{\As}{\mathsf{A}}
\safemath{\Bs}{\mathsf{B}}
\safemath{\Cs}{\mathsf{C}}
\safemath{\Ds}{\mathsf{D}}
\safemath{\Es}{\mathsf{E}}
\safemath{\Fs}{\mathsf{F}}
\safemath{\Gs}{\mathsf{G}}
\safemath{\Hs}{\mathsf{H}}
\safemath{\Is}{\mathsf{I}}
\safemath{\Js}{\mathsf{J}}
\safemath{\Ks}{\mathsf{K}}
\safemath{\Ls}{\mathsf{L}}
\safemath{\Ms}{\mathsf{M}}
\safemath{\Ns}{\mathsf{N}}
\safemath{\Os}{\mathsf{O}}
\safemath{\Ps}{\mathsf{P}}
\safemath{\Qs}{\mathsf{Q}}
\safemath{\Rs}{\mathsf{R}}
\safemath{\Ss}{\mathsf{S}}
\safemath{\Ts}{\mathsf{T}}
\safemath{\Us}{\mathsf{U}}
\safemath{\Vs}{\mathsf{V}}
\safemath{\Ws}{\mathsf{W}}
\safemath{\Xs}{\mathsf{X}}
\safemath{\Ys}{\mathsf{Y}}
\safemath{\Zs}{\mathsf{Z}}

\safemath{\as}{\mathsf{a}}
\safemath{\bs}{\mathsf{b}}
\safemath{\cs}{\mathsf{c}}
\safemath{\ds}{\mathsf{d}}
\safemath{\es}{\mathsf{e}}
\safemath{\fs}{\mathsf{f}}
\safemath{\gs}{\mathsf{g}}
\safemath{\hs}{\mathsf{h}}
\safemath{\is}{\mathsf{i}}
\safemath{\js}{\mathsf{j}}
\safemath{\ks}{\mathsf{k}}
\safemath{\ls}{\mathsf{l}}
\safemath{\ms}{\mathsf{m}}
\safemath{\ns}{\mathsf{n}}
\safemath{\os}{\mathsf{o}}
\safemath{\ps}{\mathsf{p}}
\safemath{\qs}{\mathsf{q}}
\safemath{\rs}{\mathsf{r}}
\safemath{\us}{\mathsf{u}}
\safemath{\xs}{\mathsf{x}}
\safemath{\ys}{\mathsf{y}}
\safemath{\zs}{\mathsf{z}}

\DeclareMathAlphabet{\mathbfsf}{\encodingdefault}{\sfdefault}{bx}{n}
\safemath{\Asf}{\mathbfsf{A}}
\safemath{\Bsf}{\mathbfsf{B}}
\safemath{\Csf}{\mathbfsf{C}}
\safemath{\Dsf}{\mathbfsf{D}}
\safemath{\Esf}{\mathbfsf{E}}
\safemath{\Fsf}{\mathbfsf{F}}
\safemath{\Gsf}{\mathbfsf{G}}
\safemath{\Hsf}{\mathbfsf{H}}
\safemath{\Isf}{\mathbfsf{I}}
\safemath{\Jsf}{\mathbfsf{J}}
\safemath{\Ksf}{\mathbfsf{K}}
\safemath{\Lsf}{\mathbfsf{L}}
\safemath{\Msf}{\mathbfsf{M}}
\safemath{\Nsf}{\mathbfsf{N}}
\safemath{\Osf}{\mathbfsf{O}}
\safemath{\Psf}{\mathbfsf{P}}
\safemath{\Qsf}{\mathbfsf{Q}}
\safemath{\Rsf}{\mathbfsf{R}}
\safemath{\Ssf}{\mathbfsf{S}}
\safemath{\Tsf}{\mathbfsf{T}}
\safemath{\Usf}{\mathbfsf{U}}
\safemath{\Vsf}{\mathbfsf{V}}
\safemath{\Wsf}{\mathbfsf{W}}
\safemath{\Xsf}{\mathbfsf{X}}
\safemath{\Ysf}{\mathbfsf{Y}}
\safemath{\Zsf}{\mathbfsf{Z}}

\safemath{\asf}{\mathbfsf{a}}
\safemath{\bsf}{\mathbfsf{b}}
\safemath{\csf}{\mathbfsf{c}}
\safemath{\dsf}{\mathbfsf{d}}
\safemath{\esf}{\mathbfsf{e}}
\safemath{\fsf}{\mathbfsf{f}}
\safemath{\gsf}{\mathbfsf{g}}
\safemath{\hsf}{\mathbfsf{h}}
\safemath{\isf}{\mathbfsf{i}}
\safemath{\jsf}{\mathbfsf{j}}
\safemath{\ksf}{\mathbfsf{k}}
\safemath{\lsf}{\mathbfsf{l}}
\safemath{\msf}{\mathbfsf{m}}
\safemath{\nsf}{\mathbfsf{n}}
\safemath{\osf}{\mathbfsf{o}}
\safemath{\psf}{\mathbfsf{p}}
\safemath{\qsf}{\mathbfsf{q}}
\safemath{\rsf}{\mathbfsf{r}}
\safemath{\ssf}{\mathbfsf{s}}
\safemath{\tsf}{\mathbfsf{t}}
\safemath{\usf}{\mathbfsf{u}}
\safemath{\vsf}{\mathbfsf{v}}
\safemath{\wsf}{\mathbfsf{w}}
\safemath{\xsf}{\mathbfsf{x}}
\safemath{\ysf}{\mathbfsf{y}}
\safemath{\zsf}{\mathbfsf{z}}

\usepackage{amsthm}
\newtheorem{problem}{Problem}
\newtheorem{theorem}{Theorem}
\newtheorem{lemma}{Lemma}
\newtheorem{strategy}{Strategy}

\title{Optimal transport maps for distribution preserving operations on latent spaces of Generative Models}

\author{Eirikur Agustsson\\ 
D-ITET, ETH Zurich\\
Switzerland\\
\texttt{aeirikur@vision.ee.ethz.ch} \\
\And
Alexander Sage\\ 
D-ITET, ETH Zurich\\
Switzerland\\
\texttt{sagea@student.ethz.ch~~~~~~~~} \\
\And 
Radu Timofte\\ 
D-ITET, ETH Zurich\\
Merantix GmbH\\
\texttt{radu.timofte@vision.ee.ethz.ch} \\
\And 
Luc Van Gool\\ 
D-ITET, ETH Zurich\\
ESAT, KU Leuven\\
\texttt{vangool@vision.ee.ethz.ch} \\
}

\iclrfinalcopy 

\begin{document}

\maketitle

\vspace{-0.2cm}
\begin{abstract}
\vspace{-0.2cm}
Generative models such as Variational Auto Encoders (VAEs) and Generative Adversarial Networks (GANs) are typically trained for a fixed prior distribution in the latent space, such as uniform or Gaussian.
After a trained model is obtained, one can sample the Generator in various forms for exploration and understanding, such as interpolating between two samples, sampling in the vicinity of a sample or exploring differences between a pair of samples applied to a third sample.
In this paper, we show that the latent space operations used in the literature so far induce a distribution mismatch between the resulting outputs and the prior distribution the model was trained on. To address this, we propose to use distribution matching transport maps to ensure that such  latent space operations preserve the prior distribution, while minimally modifying the original operation. 
Our experimental results validate that the proposed operations give higher quality samples compared to the original operations.
\end{abstract}

\vspace{-0.2cm}
\section{Introduction \& Related Work}
Generative models such as Variational Autoencoders (VAEs)~\citep{VAE} and Generative Adversarial Networks (GANs)~\citep{goodfellow2014generative} have emerged as popular techniques for unsupervised learning of intractable distributions.
In the framework of Generative Adversarial Networks (GANs)~\citep{goodfellow2014generative}, the generative model is obtained by jointly training a generator $G$ and a discriminator $D$ in an adversarial manner. The discriminator is trained to classify synthetic samples from real ones, whereas the generator is trained to map samples drawn from a fixed prior distribution to synthetic examples which fool the discriminator.
Variational Autoencoders (VAEs)~\citep{VAE} are also trained for a fixed prior distribution, but this is done through the loss of an Autoencoder that minimizes the variational lower bound of the data likelihood.
For both VAEs and GANs, using some data $\Xc$ we end up with a trained generator $G$, that is supposed to map latent samples $\z$ from the fixed prior distribution to output samples $G(\z)$ which (hopefully) have the same distribution as the data.

In order to understand and visualize the learned model $G(z)$, it is a common practice in the literature of generative models to explore how the output $G(z)$ behaves under various arithmetic operations on the  latent samples $z$.
In this paper, we show that the operations typically used so far, such as linear interpolation~\citep{goodfellow2014generative}, spherical interpolation~\citep{white2016sampling}, vicinity sampling and vector arithmetic~\citep{DCGAN}, cause a distribution mismatch between the latent prior distribution and the results of the operations.
This is problematic, since the generator $G$ was trained on a fixed prior and expects to see inputs with statistics consistent with that distribution. We show that this, somewhat paradoxically, is also a problem if the support of resulting (mismatched) distribution is within the support of a uniformly distributed prior, whose points all have equal likelihood during training.

\begin{figure}[ht!]
\vspace{-1.3cm}
  \centering
  \resizebox{\linewidth}{!}
  {
  \begin{tabular}{ccc}
    \multirow{1}{0.55\linewidth}
    {
    \begin{subfigure}[t]{0.99\linewidth}
    \vspace{-2.2cm}
    \includegraphics[width=\linewidth]{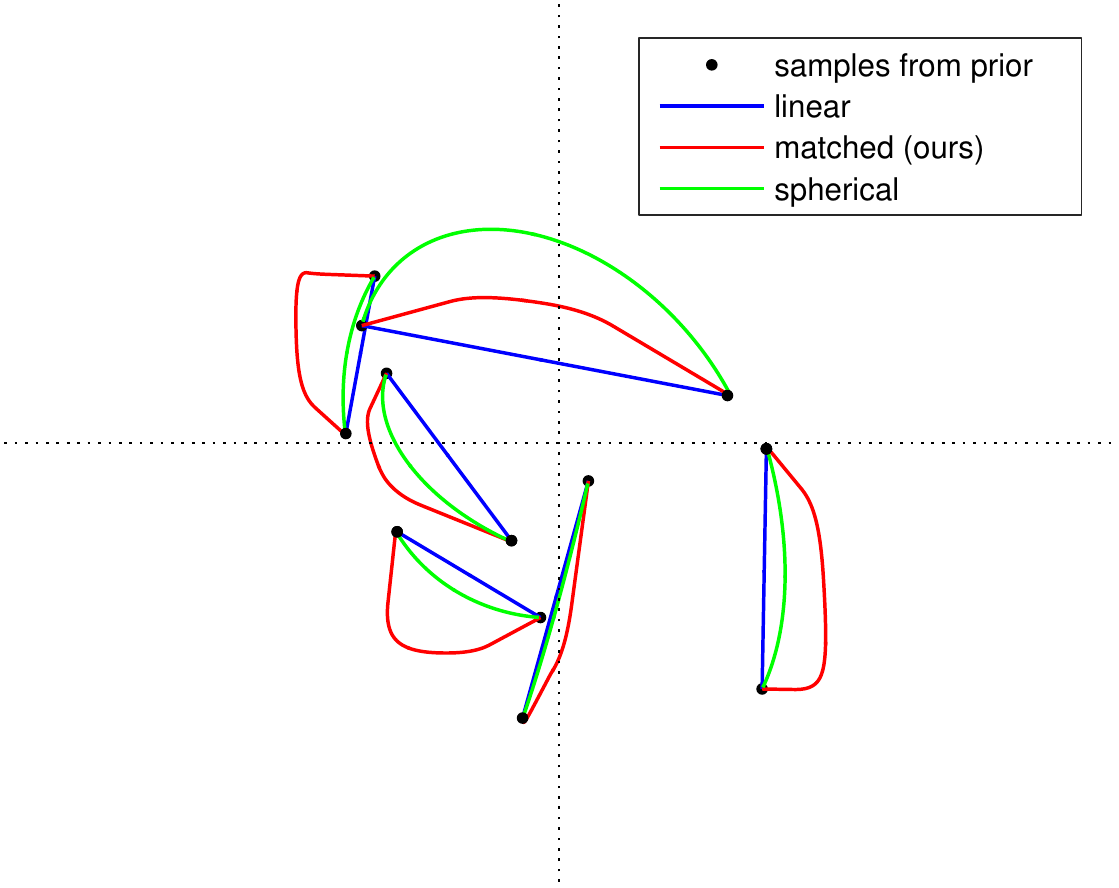}
    \caption{\label{fig:traj} \textbf{Uniform prior:} Trajectories of linear interpolation, our matched interpolation and the spherical interpolation~\citep{white2016sampling}.}
  \end{subfigure}
  } 
  & 
  \begin{subfigure}[c]{0.24\linewidth}
    \includegraphics[width=0.95\linewidth]{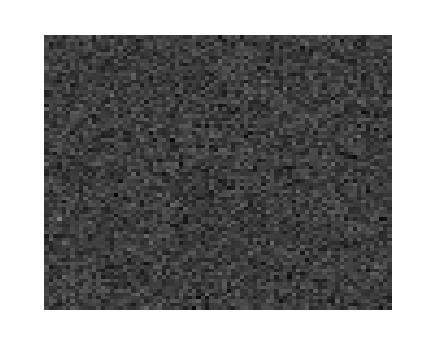}
    \caption{\label{fig:prior} Uniform prior distribution.}
  \end{subfigure}%
  & \begin{subfigure}[c]{0.24\linewidth}
    \includegraphics[width=0.95\linewidth]{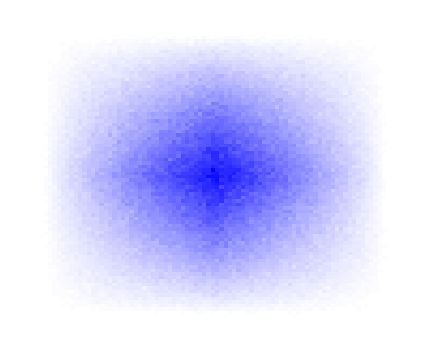}
    \caption{\label{fig:linear} Linear midpoint distribution}
  \end{subfigure} \\
   & \begin{subfigure}[c]{0.24\linewidth}
    \includegraphics[width=0.95\linewidth]{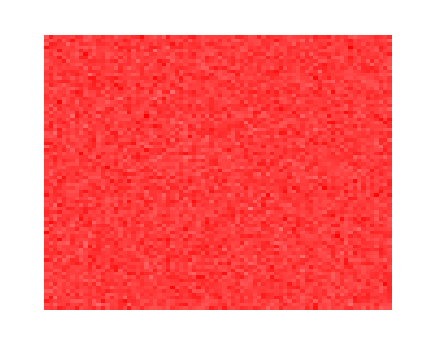}
    \caption{\label{fig:matched} Matched midpoint distribution (\textbf{ours})}
  \end{subfigure}
  &
   \begin{subfigure}[c]{0.24\linewidth}
    \centering\includegraphics[width=0.95\linewidth]{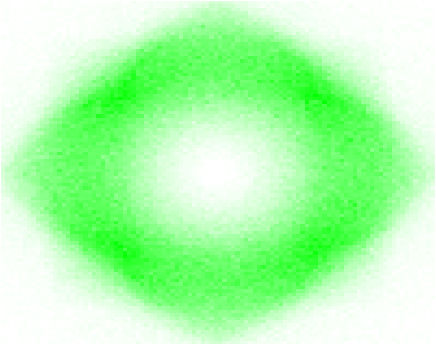}
    \caption{\label{fig:spherical} Spherical midpoint distribution~\citep{white2016sampling}}
  \end{subfigure}\\
  \end{tabular}
  }
  \vspace{0.3cm}
  \caption{\label{fig:teaser}
  We show examples of distribution mismatches induced by the previous interpolation schemes when using a uniform prior in two dimensions.
Our matched interpolation avoids this with a minimal modification to the linear trajectory, traversing through the space such that all points along the path are distributed identically to the prior.}
\end{figure}

To address this, we propose to use distribution matching transport maps, to obtain analogous latent space operations (e.g. interpolation, vicinity sampling) which preserve the prior distribution of the latent space, while minimally changing the original operation.
In Figure~\ref{fig:teaser} we showcase how our proposed technique gives an interpolation operator which avoids distribution mismatch when interpolating between samples of a uniform distribution. The points of the (red) matched trajectories are obtained as minimal deviations (in expectation of $l_1$ distance) from the the points of the (blue) linear trajectory.

\subsection{Generative Models and Sample Operations}
\label{sec:intro_operators}
In the literature there are dozens of papers that use sample operations to explore the learned models.
\cite{bengio2013better} use linear interpolation between neighbors in the latent space to study how well deep vs shallow representations can disentangle the latent space of Contractive Auto Encoders (CAEs)~\citep{CAE}.

In the seminal GAN paper of \cite{goodfellow2014generative}, the authors use linear interpolation between latent samples to visualize the transition between outputs of a GAN trained on MNIST.
\cite{dosovitskiy2015learning} linearly interpolate the latent codes of an auto encoder trained on a synthetic chair dataset.

\cite{DCGAN} also linearly interpolate between samples to evaluate the quality of the learned representation. Furthermore, motivated by the semantic word vectors of~\cite{mikolov2013distributed}, they explore using vector arithmetic on the samples to change semantics such as adding a smile to a generated face.

\cite{reed2016generative} use linear interpolation to explore their proposed GAN model which operates jointly in the visual and textual domain.
\cite{brock2016neural} combine GANs and VAEs for a neural photo editor, using masked interpolations to edit an embedded photo in the latent space.

\subsection{Distribution Mismatch and Related Approaches}
While there are numerous works performing operations on samples, most of them have ignored the problem of distribution mismatch, such as the one presented in Figure~\ref{fig:matched}.
\cite{VAE} and \cite{makhzani2015adversarial} sidestep the problem when visualizing their models, by not performing operations on latent samples, but instead restrict the latent space to 2-d and uniformly sample the percentiles of the distribution on a 2-d grid. This way, the samples have statistics that are consistent with the prior distribution. However, this approach does not scale up to higher dimensions - whereas the latent spaces used in the literature can have hundreds of dimensions.

Related to our work, \cite{white2016sampling} experimentally observe that there is a distribution mismatch between the distance to origin for points drawn from uniform or Gaussian distribution and points obtained with linear interpolation, and propose to use a so-called \textit{spherical linear interpolation} to reduce the mismatch, obtaining higher quality interpolated samples. However, the proposed approach has no theoretical guarantees.

In this work, we propose a generic method to fully preserve the desired prior distribution when using sample operations. The approach works as follows: we are given a `desired' operation, such as linear interpolation $\y= t \z_1 + (1-t)\z_2$, $t\in[0,1]$. Since the distribution of $\y$ does not match the prior distribution of $\z$, we search for a warping $f:\R^d\to \R^d$, such that $\tilde{\y}=f(\y)$ has the same distribution as $\z$.
In order to have the modification $\tilde{\y}$ as faithful as possible to the original operation $\y$, we use optimal transform maps~\citep{santambrogio2015optimal,villani2003topics,villani2008optimal} to find a minimal modification of $\y$ which recovers the prior distribution $\z$.

This is illustrated in Figure~\ref{fig:traj}, where each point $\tilde{\y}$ of the matched curve is obtained by warping a corresponding point $\y$ of the linear trajectory, while not deviating too far from the line.

\section{From distribution mismatch to optimal transport}
\begin{table}
\vspace{-1.2cm}
\centering
\begin{tabular}{ll|l}
Operation & Expression & (Gaussian) Matched Operation \\ \hline
& & \vspace{-0.2cm}\\
\textit{{2-point interpolation}} & $\y = t \z_1 + (1-t)\z_2$ , $t \in [0,1]$ & $\tilde{\y}={\y}/{\sqrt{t^2+(1-t)^2}}$ \\
\textit{{$n$-point interpolation}} & $\y = \sum_{i=1}^n t_i \z_i$ with $\sum_i t_i = 1$ & $\tilde{\y} = \y / \sqrt{\sum_{i=1}^n t_i^2}$\\
\textit{{Vicinity sampling}} & $\y_{j} = \z_1 + \epsilon \u_j$ for $j=1,\cdots,k$ & $\tilde{\y_j}=\y_j/\sqrt{1+\epsilon^2}$\\
\textit{{Analogies}} & $\y = \z_3 + (\z_2-\z_1)$ & $\tilde{\y}=\y/\sqrt{3}$ \\
\end{tabular}
\caption{Examples of interesting sample operations which need to be adapted if we want the distribution of the result $\y$ to match the prior distribution. If the prior is Gaussian, our proposed matched operation simplifies to a proper re-scaling factor (see third column) for additive operations.
}
\label{tab:operations}
\end{table}

With implicit models such as GANs~\citep{goodfellow2014generative} and VAEs~\citep{VAE}, we use the data $\Xc$, drawn from an unknown random variable $\x$, to learn a generator $G:\R^d \mapsto \R^{d'}$ with respect to a fixed prior distribution $p_{\z}$, such that $G(\z)$ approximates $\x$. Once the model is trained, we can sample from it by feeding latent samples $\z$ through $G$.

We now bring our attention to \textit{operations} on latent samples $\z_1,\cdots,\z_k$ from $p_{\z}$, i.e. mappings
\begin{equation}
        \kappa: \R^d\times \cdots \times \R^d \to \R^d.
\end{equation}
We give a few examples of such operations in Table~\ref{tab:operations}.

Since the inputs to the operations are random variables, their output $\y=\kappa(\z_1,\cdots,\z_k)$ is also a random variable (commonly referred to as a \textit{statistic}).
While we typically perform these operations on \textit{realized} (i.e. observed) samples, our analysis is done through the underlying random variable $\y$. The same treatment is typically used to analyze other statistics over random variables, such as the sample mean, sample variance and test statistics.

In Table~\ref{tab:operations} we show example operations which have been commonly used in the literature. As discussed in the Introduction, such operations can provide valuable insight into how the trained generator $G$ changes as one creates related samples $\y$ from some source samples.
The most common such operation is the linear interpolation, which we can view as an operation
\begin{equation}
\y_t = t\z_1 + (1-t)\z_2,
\end{equation}
where $\z_1,\z_2$ are latent samples from the prior $p_{\z}$ and $\y_t$ is parameterized by $t\in[0,1]$.

Now, assume $\z_1$ and $\z_2$ are i.i.d, and let $Z_1,Z_2$ be their (scalar) first components with distribution $p_{Z}$.
Then the first component of $\y_t$ is $Y_t = tZ_1 + (1-t)Z_2$, and we can compute:
\begin{equation}
        \text{Var}[Y_{t}] = \text{Var}[tZ_1 + (1-t) Z_2] = t^2 \text{Var}[Z_1] +  (1-t)^2 \text{Var}[Z_2] = (1+2t(t-1))\text{Var}[Z].
\end{equation}
Since $(1+2t(t-1)) \neq 1$ for all $t\in [0,1]\setminus \{0,1\}$, it is in general impossible for $\y_t$ to have the same distribution as $\z$, which means that distribution mismatch is \textit{inevitable} when using linear interpolation. A similar analysis reveals the same for all of the operations in Table~\ref{tab:operations}.

This leaves us with a dilemma: we have various intuitive operations (see Table~\ref{tab:operations}) which we would want to be able to perform on samples, but their resulting distribution $p_{\y_t}$ is inconsistent with the distribution $p_{\z}$ we trained $G$ for. 

\begin{figure}[tb]
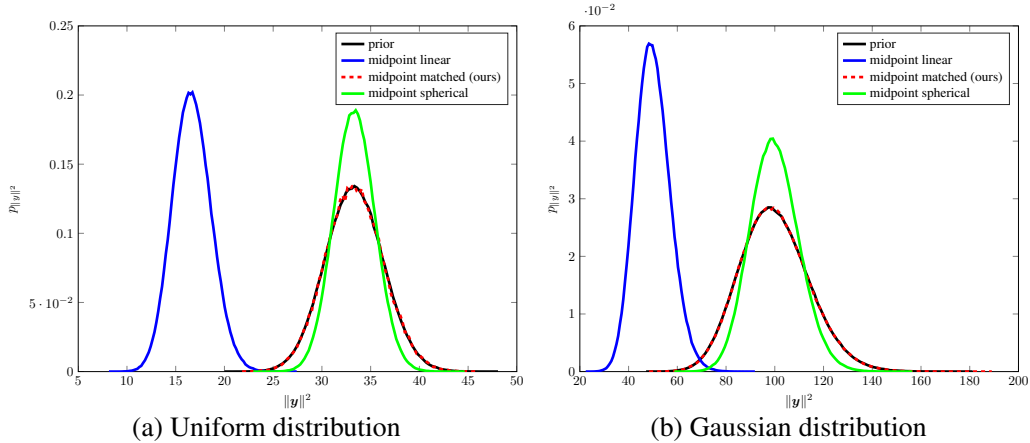

\vspace{-1.3cm}
  \centering
  \resizebox{\linewidth}{!}
  {
  \begin{tabular}{cc}
  \input{squared_norm_distr.tex}&
  \input{squared_norm_distr_gauss.tex}\\
  \huge (a) Uniform distribution & \huge (b) Gaussian distribution\\
  \end{tabular}
  }
  \caption{
  \label{fig:squared_norm_distr}
  Distribution of the squared norm $\|\y\|^2$ of midpoints for two prior distributions in $100$ dimensions: (a) components uniform on $[-1,1]$ and (b) components Gaussian $\mathcal{N}(0,1)$, for linear interpolation, our proposed matched interpolation and the spherical interpolation proposed by \cite{white2016sampling}.
Both linear and spherical interpolation introduce a distribution mismatch, whereas our proposed matched interpolation preserves the prior distribution for both priors. 
}
\end{figure}

Due to the \textit{curse of dimensionality}, as empirically observed by \cite{white2016sampling}, this mismatch can be significant in high dimensions. 
We illustrate this in Figure~\ref{fig:squared_norm_distr}, where we plot the distribution of the squared norm $\|\y_t\|^2$ for the midpoint $t=1/2$ of linear interpolation, compared to the prior distribution $\|\z\|^2$. With $d=100$ (a typical dimensionality for the latent space), the distributions are dramatically different, having almost no common support. In Appendix~\ref{app:curse} we expand this analysis and show that  this happens for all prior distributions with i.i.d. entries (i.e. not only Uniform and Gaussian).


\subsection{Distribution matching with optimal transport}

In order to address the distribution mismatch, we propose a simple and intuitive strategy for constructing distribution preserving operators, via optimal transport:
\begin{strategy}[Optimal Transport Matched Operations]\label{transport_strategy}
\quad

\begin{enumerate}
\item We construct an 'intuitive' operator $\y=\kappa(\z_1,\cdots,\z_k)$.
\item We analytically (or numerically) compute the resulting (mismatched) distribution $p_{\y}$
\item\label{transport} We search for a minimal modification $\tilde{\y}=f(\y)$ (in the sense that $E_{\y}[c(\tilde{\y},\y)]$ is minimal with respect to a cost $c$), such that distribution is brought back to the prior, i.e. $p_{\tilde{\y}}=p_{\z}$.
\end{enumerate}
\end{strategy}
The cost function in step \ref{transport} could e.g. be the euclidean distance $c(x,y)=\|x-y\|$, and is used to measure how faithful the modified operator, $\tilde{\y}=f(\kappa(\z_1,\cdots,\z_k))$ is to the original operator $k$. Finding the map $f$ which gives a minimal modification can be challenging, but fortunately it is a well studied problem from optimal transport theory.
We refer to the modified operation $\tilde{y}$ as the \textit{matched} version of $\y$, with respect to the cost $c$ and prior distribution $p_{\z}$.

For completeness, we introduce the key concept of optimal transport theory in a simplified setting, i.e. assuming probability distributions are in euclidean space and skipping measure theoretical formalism. We refer to \cite{villani2003topics,villani2008optimal} and \cite{santambrogio2015optimal} for a thorough and formal treatment of optimal transport.

The problem of step \eqref{transport} above was first posed by \cite{monge1781memoire} and can more formally be stated as:
\begin{problem}[\cite{santambrogio2015optimal} Problem 1.1]
\label{monge}
Given probability distributions $p_{\x},p_{\y}$, with domains $\Xc,\Yc$ respectively, and a cost function $c:\Xc\times\Yc\to\R^+$, we want to minimize
\begin{equation}
inf \left\{ E_{\x\sim p_{\x}}[c(\x,f(\x))] \Big| f:\Xc\to\Yc, f(\x)\sim p_{\y} \right\} \tag{MP} \label{eq:MP}
\end{equation}
We refer to the minimizer $f^* \Xc\to\Yc$ of \eqref{eq:MP} (if it exists), as the optimal transport map from $p_{\x}$ to $p_{\y}$ with respect to the cost $c$.
\end{problem}
However, the problem remained unsolved until a relaxed problem was studied by~\cite{kantorovich1942translocation}:

\begin{problem}[\cite{santambrogio2015optimal} Problem 1.2]
\label{kantorovich}
Given probability distributions $p_{\x},p_{\y}$, with domains $\Xc,\Yc$ respectively, and a cost function $c:\Xc\times\Yc\to\R^+$, we want to minimize
\begin{equation}
inf \left\{ E_{(\x,\y)\sim p_{\x,\y}}[c(\x,\y)] \Big|  (\x,\y)\sim p_{\x,\y}, \x\sim p_{\x}, \y\sim p_{\y} \right\} \tag{KP}\label{eq:KP},
\end{equation}
where $(\x,\y)\sim p_{\x,\y}, \x\sim p_{\x}, \y\sim p_{\y}$ denotes that $(\x,\y)$ have a joint distribution $p_{\x,\y}$ which has (previously specified) marginals $p_{\x}$ and $p_{\y}$.

We refer to the joint $p_{\x,\y}$ which minimizes \eqref{eq:KP} as the optimal transport plan from $p_{\x}$ to $p_{\y}$ with respect to the cost $c$.
\end{problem}

The key difference is to relax the deterministic relationship between $\x$ and $f(\x)$ to a joint probability distribution $p_{\x,\y}$ with marginals $p_{\x}$ and $p_{\y}$ for $\x$ and $\y$. In the case of Problem~\ref{monge}, the minimization might be over the empty set since it is not guaranteed that there exists a mapping $f$ such that $f(\x)\sim\y$. In contrast, for Problem~\ref{kantorovich}, one can always construct a joint density $p_{\x,\y}$ with $p_{\x}$ and $p_{\y}$ as marginals, such as the trivial construction where $\x$ and $\y$ are independent, i.e. $p_{\x,\y}(x,y):=p_{\x}(x)p_{\y}(y)$.

Note that given a joint density $p_{\x,\y}(x,y)$ over $\Xc\times \Yc$, we can view $\y$ conditioned on $\x=x$ for a fixed $x$ as a stochastic function $\f(x)$ from $\Xc$ to $\Yc$, since given a fixed $x$ do not get a specific function value $f(x)$ but instead a random variable $\f(x)$ that depends on $x$, with $\f(x) \sim \y | \x=x$ with density $p_{\y}(y|\x=x):=\frac{p_{\x,\y}(x,y)}{p_{\x}(x)}$.
In this case we have $(\x,\f(\x)) \sim p_{\x,\y}$, so we can view the Problem~\ref{eq:KP} as a relaxation of Problem~\ref{eq:MP} where $f$ is allowed to be a stochastic mapping.

While the relaxed problem of Kantorovich \eqref{eq:KP} is much more studied in the optimal transport literature, for our purposes of constructing operators it is desirable for the mapping $f$ to be deterministic as in \eqref{eq:MP}.

To this end, we will choose the cost function $c$ such that the two problems coincide and where we can find an analytical solution $f$ or at least an efficient numerical solution.

In particular, we note that most operators in Table~\ref{tab:operations} are all \textit{pointwise}, such that if the points $\z_i$ have i.i.d. components, then the result $\y$ will also have i.i.d. components.

If we combine this with the constraint for the cost $c$ to be additive over the components of $\x,\y$, we obtain the following simplification:
\begin{theorem}
        \label{1dthm}
        Suppose $p_{\x}$ and $p_{\y}$ have i.i.d components and $c$ over $\Xc\times \Yc=\R^d\times \R^d$ decomposes as
\begin{equation}
        c(x,y) = \sum_{i=1}^d  C(x^{(i)},y^{(i)})\label{sumcost}.
\end{equation}
Consequently, the minimization problems \eqref{eq:MP} and \eqref{eq:KP} turn into $d$ identical scalar problems for the distributions $p_X$ and $p_Y$ of the components of $\x$ and $\y$:
\begin{align}
        inf \left\{ E_{X\sim p_{X}}[C(X,T(X))] \Big| T:\R\to\R, T(X)\sim p_{Y} \right\} \tag{MP-1-D}\label{eq:MP-1-D}\\
        inf \left\{ E_{(X,Y)\sim p_{X,Y}}[C(X,Y)] \Big|  (X,Y)\sim p_{X,Y}, X\sim p_{X}, Y\sim p_{Y} \right\} \tag{KP-1-D}\label{eq:KP-1-D},
\end{align}
such that an optimal transport map $T$ for \eqref{eq:MP-1-D} gives an optimal transport map $f$ for \eqref{eq:MP} by pointwise application of $T$, i.e. $f(x)^{(i)}:=T(x^{(i)})$,
and an optimal transport plan $p_{X,Y}$ for \eqref{eq:KP-1-D} gives an optimal transport plan $p_{\x,\y}(x,y):=\prod_{i=1}^dp_{X,Y}(x^{(i)},y^{(i)})$ for \eqref{eq:KP}.
        \begin{proof}
                See Appendix.
        \end{proof}
\end{theorem}

Fortunately, under some mild constraints, the scalar problems have a known solution:
\begin{theorem}[Theorem 2.9 in \cite{santambrogio2015optimal}]\label{monotone_optimal}
       Let $h:\R\to \R^+$ be convex and suppose the cost $C$ takes the form $C(x,y)=h(x-y)$.
       Given an continuous source distribution $p_{X}$ and a target distribution $p_{Y}$ on $\R$ having a finite optimal transport cost in \eqref{eq:KP-1-D}, then
       \begin{equation}
               T_{X\to Y}^{\text{mon}}(x):= F^{[-1]}_Y(F_X(x)),\label{monotone_map}
       \end{equation}
       defines an optimal transport map from $p_X$ to $p_Y$ for \eqref{eq:MP-1-D},
       where $F_X(x):=\int_{-\infty}^x p_X(x') dx'$ is the Cumulative Distribution Function (CDF) of $X$ and $F^{[-1]}_Y(y):= \inf\{t\in \R | F_Y(t)\geq y\}$ is the pseudo-inverse of $F_Y$.
       Furthermore, the joint distribution of $(X,T_{X\to Y}^{\text{mon}}(X))$ defines an optimal transport plan for \eqref{eq:KP-1-D}.
\end{theorem}
The mapping $T_{X\to Y}^{\text{mon}}(x)$ in Theorem~\ref{monotone_optimal} is non-decreasing and is known as the \textit{monotone transport map} from $X$ to $Y$.
It is easy to verify that  $T_{X\to Y}^{\text{mon}}(X)$  has the distribution of $Y$, in particular $F_X(X)\sim \text{Uniform}(0,1)$ and if $U\sim \text{Uniform}(0,1)$ then $F^{[-1]}_Y(U)\sim Y$.

Now, combining Theorems~\ref{1dthm} and \ref{monotone_optimal}, we obtain a concrete realization of the Strategy~\ref{transport_strategy} outlined above. We choose the cost $c$ such that it admits to Theorem~\ref{1dthm}, such as $c(\x,\y):=\|\x-\y\|_1$, and use an operation that is pointwise, so we just need to compute the monotone transport map in \eqref{monotone_map}.
That is, if $\z$ has i.i.d components with  distribution $p_Z$, we just need to compute the component distribution $p_{Y}$ of the result $\y$ of the operation, the CDFs $F_Z,F_Y$ and obtain
\begin{equation}
    T_{Y\to Z}^{\text{mon}}(y):= F^{[-1]}_Z(F_Y(y))
\end{equation}
as the component-wise modification of $\y$, i.e. $\tilde{\y}^{(i)}:=T_{Y\to Z}^{\text{mon}}(\y^{(i)})$.

\begin{figure}[tb]
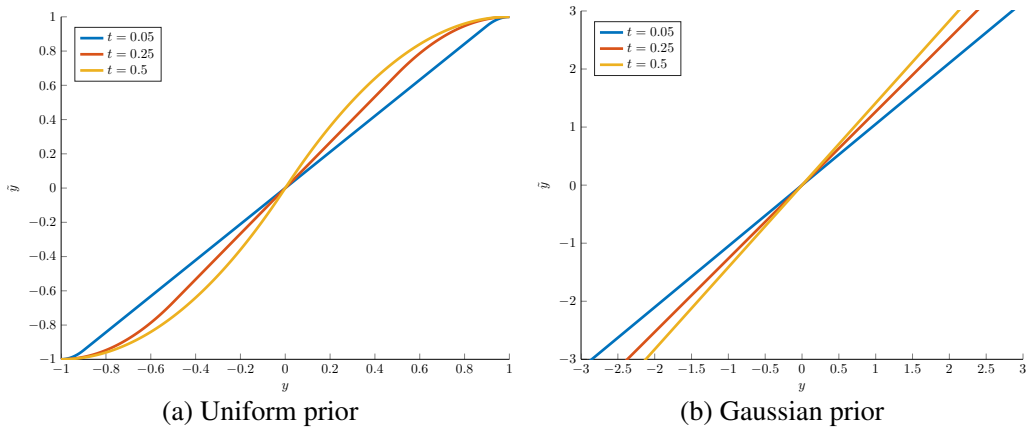

        \vspace{-1.25cm}
  \centering
  \resizebox{\linewidth}{!}
  {
  \begin{tabular}{cc}
  \input{monotone_map_unif.tex}&
  \input{monotone_map_gauss.tex}\\
  \huge (a) Uniform prior & \huge (b) Gaussian prior \\
  \end{tabular}
  }
  \caption{
  \label{fig:transport_maps_interp}
  We show the monotone transport maps for linear interpolation evaluated at $t\in\{0.05, 0.25, 0.5\}$, to Uniform and Gaussian priors.
}
\end{figure}
In Figure~\ref{fig:transport_maps_interp} we show the monotone transport map for the linear interpolation $\y=t\z_1+(1-t)\z_2$ for various values of $t$.
The detailed calculations and examples for various operations are given in Appendix~\ref{app:examples}, for both Uniform and Gaussian priors.
The Gaussian case has a particularly simple resulting transport map for additive operations, where it is just a linear transformation through a scalar multiplication, summarized in the third column of Table~\ref{tab:operations}.

\section{Experiments}
\label{sec:experiments}

\subsection{Comparison of distributions}
To validate the correctness of the matched operators obtained above, we numerically simulate the distributions for toy examples, as well as prior distributions typically used in the literature.

\paragraph*{Priors vs. interpolations in $2$-D}
For Figure~\ref{fig:teaser}, we sample 1 million pairs of points in two dimension, from a uniform prior (on $[-1,1]^2$), and estimate numerically the midpoint distribution of linear interpolation, our proposed matched interpolation and the spherical interpolation of \cite{white2016sampling}. It is reassuring to see that the matched interpolation gives midpoints which are identically distributed to the prior. In contrast, the linear interpolation condenses more towards the origin, forming a pyramid-shaped distribution (the result of convolving two boxes in 2-d).
Since the spherical interpolation of \cite{white2016sampling} follows a great circle with varying radius between the two points, we see that the resulting distribution has a ``hole'' in it, ``circling'' around the origin for both priors. 

\paragraph*{Priors vs. interpolations in $100$-D}
For Figure~\ref{fig:squared_norm_distr}, we sample 1 million pairs of points in $d=100$ dimensions, using either i.i.d. uniform components on $[-1,1]$ or Gaussian $\mathcal{N}(0,1)$ and compute the distribution of the squared norm of the midpoints.
We see there is a dramatic difference between vector lengths in the prior and the midpoints of linear interpolation, with only minimal overlap.
We also show the spherical (SLERP) interpolation of~\cite{white2016sampling} which has a matching first moment, but otherwise also induces a distribution mismatch. In contrast, our matched interpolation, fully preserves the prior distribution and perfectly aligns. We note that this setting ($d=100$, uniform or Gaussian) is commonly used in the literature.

\subsection{Qualitative results}
\begin{figure}[tb]
\vspace{-1.25cm}
	\centering
	\begin{subfigure}[b]{0.33\linewidth}
		\centering
		\includegraphics[width=0.95\linewidth]{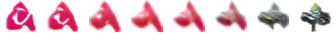}\\
		\includegraphics[width=0.95\linewidth]{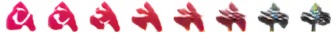}\\
		\includegraphics[width=0.95\linewidth]{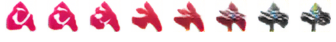}\\
		\includegraphics[width=0.95\linewidth]{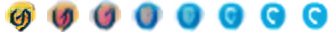}\\
		\includegraphics[width=0.95\linewidth]{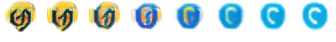}\\
		\includegraphics[width=0.95\linewidth]{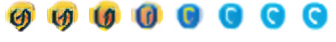}\\
		\includegraphics[width=0.95\linewidth]{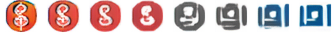}\\
		\includegraphics[width=0.95\linewidth]{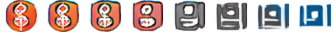}\\
		\includegraphics[width=0.95\linewidth]{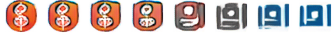}
		\caption{\label{fig:ip_icons} LLD icon dataset}
	\end{subfigure}%
	\begin{subfigure}[b]{0.33\linewidth}
		\centering
		\includegraphics[width=0.95\linewidth]{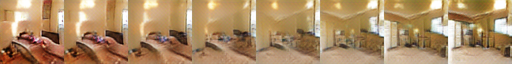}\\
		\includegraphics[width=0.95\linewidth]{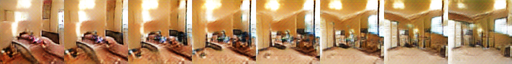}\\
		\includegraphics[width=0.95\linewidth]{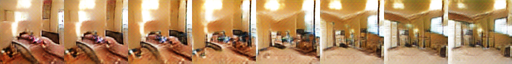}\\
		\includegraphics[width=0.95\linewidth]{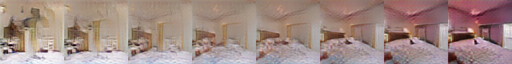}\\
		\includegraphics[width=0.95\linewidth]{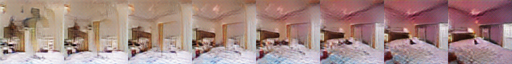}\\
		\includegraphics[width=0.95\linewidth]{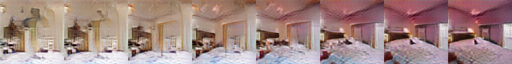}\\
		\includegraphics[width=0.95\linewidth]{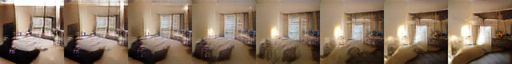}\\
		\includegraphics[width=0.95\linewidth]{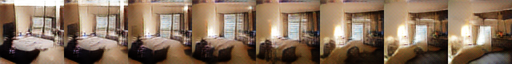}\\
		\includegraphics[width=0.95\linewidth]{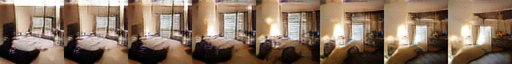}
		\caption{\label{fig:ip_lsun} LSUN dataset}
	\end{subfigure}
	\begin{subfigure}[b]{0.33\linewidth}
	\centering
	\includegraphics[width=0.95\linewidth]{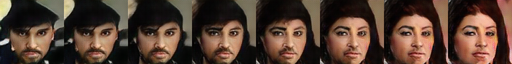}\\
	\includegraphics[width=0.95\linewidth]{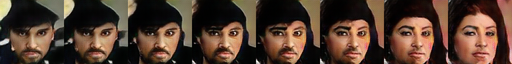}\\
	\includegraphics[width=0.95\linewidth]{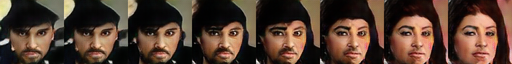}\\
	\includegraphics[width=0.95\linewidth]{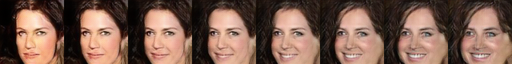}\\
	\includegraphics[width=0.95\linewidth]{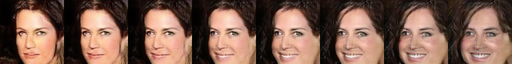}\\
	\includegraphics[width=0.95\linewidth]{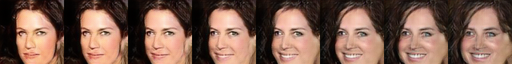}\\
	\includegraphics[width=0.95\linewidth]{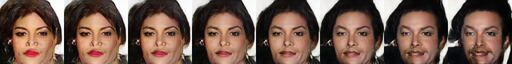}\\
	\includegraphics[width=0.95\linewidth]{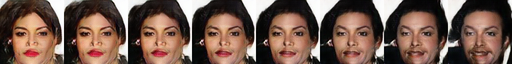}\\
	\includegraphics[width=0.95\linewidth]{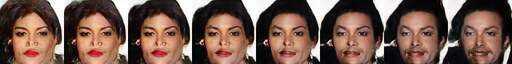}
	\caption{\label{fig:ip_celebA} CelebA dataset}
\end{subfigure}
	\caption{\label{fig:2p-interp}2-point interpolation: Each example shows linear, SLERP and transport matched interpolation from top to bottom respectively. For LLD icon dataset (a) and LSUN (b), outputs are produced with DCGAN using a uniform prior distribution, whereas the CelebA model (c) uses a Gaussian prior. The output resolution for the (a) is $32\times32$, for (b) and (c) $64\times64$ pixels.}
\end{figure}

\begin{figure}[t]
\vspace{-0.1cm}
	\centering
	\begin{subfigure}[t]{0.33\linewidth}
		\centering\includegraphics[width=0.8\linewidth]{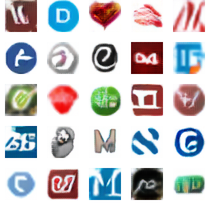}
		\caption{\label{fig:icons-midpoints-linear} Linear}
	\end{subfigure}%
	\begin{subfigure}[t]{0.33\linewidth}
		\centering\includegraphics[width=0.8\linewidth]{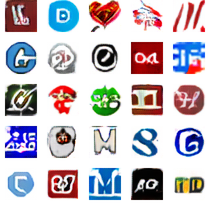}
		\caption{\label{fig:icons-midpoints-slerp} Spherical}
	\end{subfigure}
	\begin{subfigure}[t]{0.33\linewidth}
		\centering\includegraphics[width=0.8\linewidth]{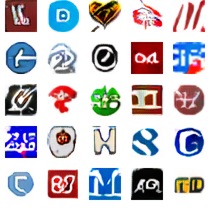}
		\caption{\label{fig:icons-midpoints-matched} Distribution matched}
	\end{subfigure}
	\caption{\label{fig:icons-midpoints}Midpoint sampling for linear, SLERP and uniform-matched interpolation when using the same pairs of sample points on LLD icon dataset with uniform prior.}
\end{figure}

\begin{figure}[t]
\vspace{-1.2cm}
	\centering
	\begin{subfigure}[t]{0.33\linewidth}
		\centering\includegraphics[width=0.95\linewidth]{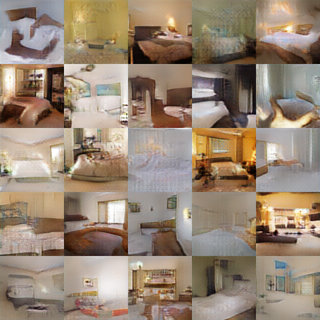}
		\caption{\label{fig:lsun-midpoints-linear} Linear}
	\end{subfigure}%
	\begin{subfigure}[t]{0.33\linewidth}
		\centering\includegraphics[width=0.95\linewidth]{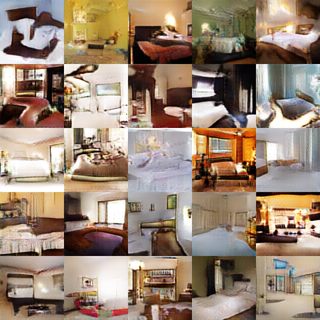}
		\caption{\label{fig:lsun-midpoints-slerp} Spherical}
	\end{subfigure}
	\begin{subfigure}[t]{0.33\linewidth}
		\centering\includegraphics[width=0.95\linewidth]{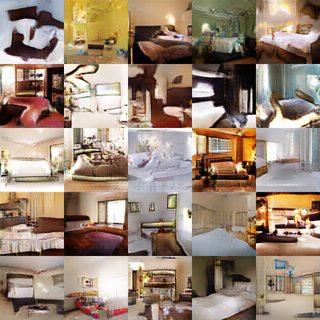}
		\caption{\label{fig:lsun-midpoints-matched} Distribution matched}
	\end{subfigure}
	\caption{\label{fig:lsun-midpoints}Midpoint sampling for linear, SLERP and uniform-matched interpolation when using the same pairs of sample points on LSUN ($64\times64$) with uniform prior.}
\end{figure}

\begin{figure}[tbh]
\vspace{-0.2cm}
	\centering
	\begin{subfigure}[t]{0.33\linewidth}
		\centering\includegraphics[width=0.95\linewidth]{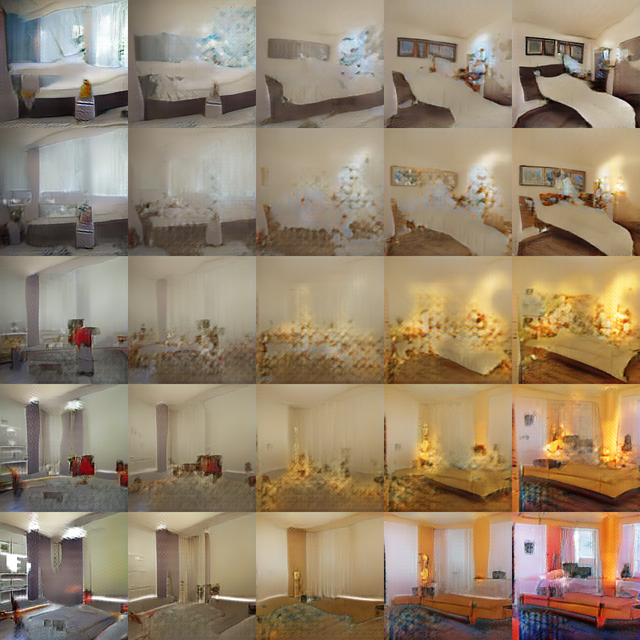}
		\caption{\label{fig:lsun-4p-linear} Linear interpolation}
	\end{subfigure}%
	\begin{subfigure}[t]{0.33\linewidth}
		\centering\includegraphics[width=0.95\linewidth]{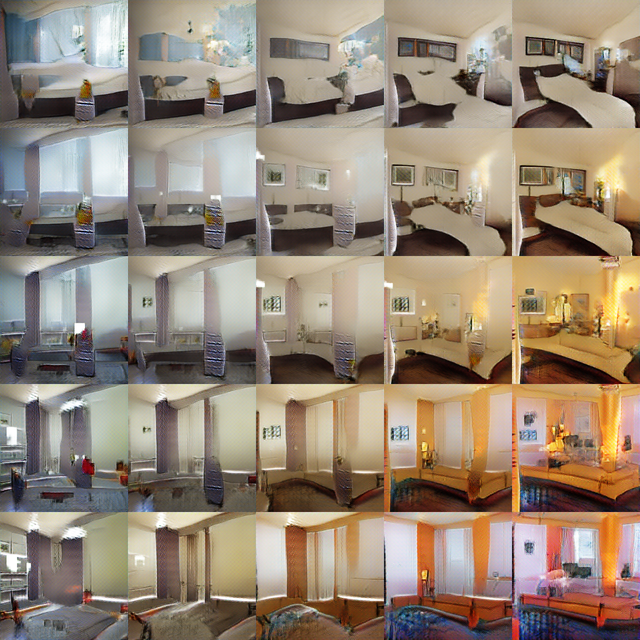}
		\caption{\label{fig:lsun-4p-slerp} Spherical interpolation}
	\end{subfigure}
	\begin{subfigure}[t]{0.33\linewidth}
		\centering\includegraphics[width=0.95\linewidth]{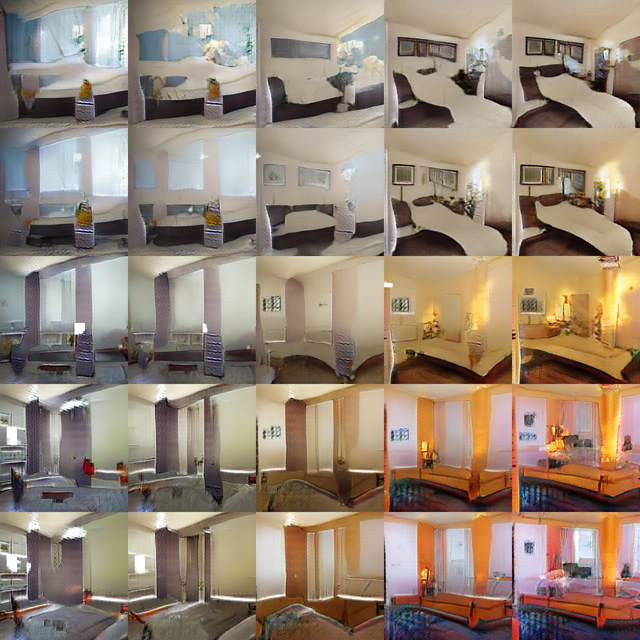}
		\caption{\label{fig:lsun-4p-matched} Distribution matched}
	\end{subfigure}
	\caption{\label{fig:lsun-4p} 4-point interpolation between 4 sampled points (corners) from DCGAN trained on LSUN ($128\times128$) using a uniform prior. The same interpolation is shown using linear, SLERP and distribution matched interpolation.}
\end{figure}

\begin{figure}[tbh]
	\centering
	\begin{subfigure}[t]{0.33\linewidth}
		\centering\includegraphics[width=0.8\linewidth]{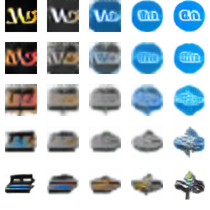}
		\caption{\label{fig:icons_4p_linear} Linear interpolation}
	\end{subfigure}%
	\begin{subfigure}[t]{0.33\linewidth}
		\centering\includegraphics[width=0.8\linewidth]{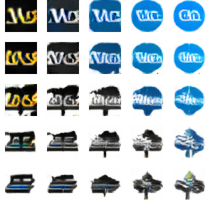}
		\caption{\label{fig:icons_4p_slerp} Spherical}
	\end{subfigure}
	\begin{subfigure}[t]{0.33\linewidth}
		\centering\includegraphics[width=0.8\linewidth]{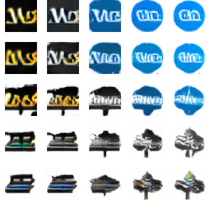}
		\caption{\label{fig:icons_4p_matched} Distribution matched}
	\end{subfigure}
	\caption{\label{fig:icons-4p} 4-point interpolation between 4 sampled points (corners) from DCGAN trained on icon dataset using a uniform prior. The same interpolation is shown using linear, SLERP and distribution matched interpolation.}
\end{figure}

\begin{figure}[tbh]
\vspace{-1.25cm}
	\centering
	\begin{subfigure}[t]{0.40\linewidth}
		\centering\includegraphics[width=0.95\linewidth]{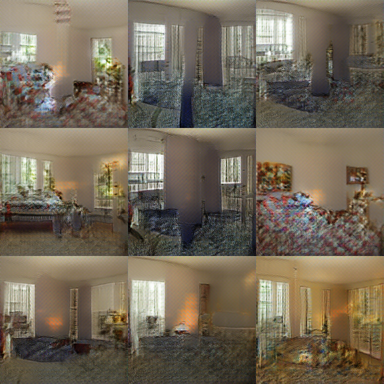}
		\caption{\label{fig:lsun_vinc_linear} Vicinity sampling}
	\end{subfigure}%
	\begin{subfigure}[t]{0.40\linewidth}
		\centering\includegraphics[width=0.95\linewidth]{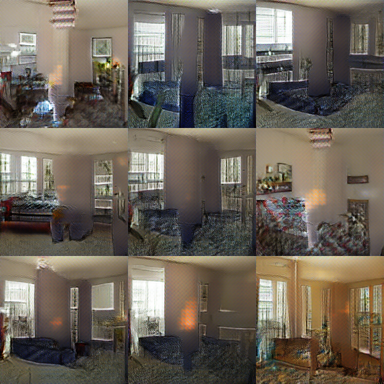}
		\caption{\label{fig:lsun_vinc_matched} Matched vicinity sampling}
	\end{subfigure}
\vspace{-0.2cm}
	\caption{\label{fig:lsun_vinc} Vicinity sampling on LSUN dataset ($128\times128$) with uniform prior. The sample in the middle is perturbed in random directions producing the surrounding sample points.}
\end{figure}

 \begin{figure}[tbh]
 	\centering
 	\begin{subfigure}[t]{0.40\linewidth}
 		\centering\includegraphics[width=0.66\linewidth]{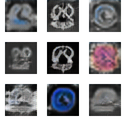}
 		\caption{\label{fig:icons_vinc_linear} Vicinity sampling}
 	\end{subfigure}%
 	\begin{subfigure}[t]{0.40\linewidth}
 		\centering\includegraphics[width=0.66\linewidth]{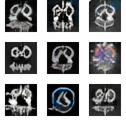}
 		\caption{\label{fig:icons_vinc_matched} Matched vicinity sampling}
 	\end{subfigure}
\vspace{-0.2cm}
 	\caption{\label{fig:icons_vinc} Vicinity sampling on LLD icon dataset with uniform prior. The sample in the middle is perturbed in random directions producing the surrounding sample points.}	
 \end{figure}

\begin{figure}[tbh]
	\centering
		\includegraphics[width=\linewidth]{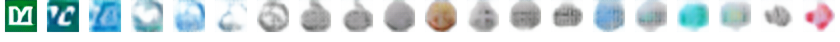}\\
	\includegraphics[width=\linewidth]{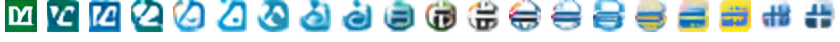}\\
	\includegraphics[width=\linewidth]{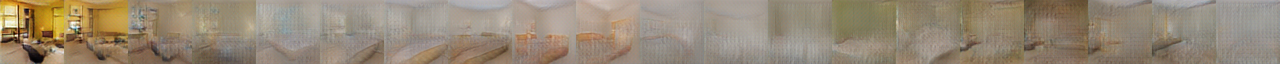}\\
	\includegraphics[width=\linewidth]{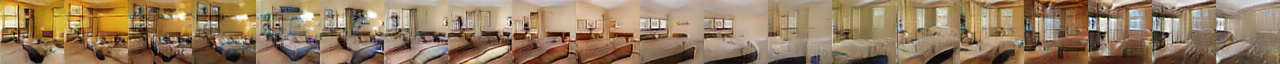}\\
	\vspace*{1ex}
	\includegraphics[width=\linewidth]{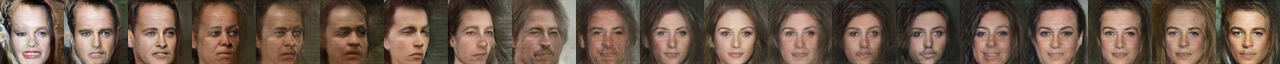}\\
	\includegraphics[width=\linewidth]{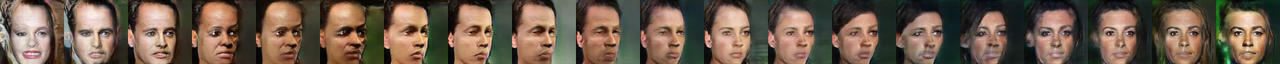}\\
	\caption{\label{fig:random-walk}Random walk for LLD, LSUN (64 x 64) and CelebA.
The random walks consist of a succession of steps in random directions, calculated for the same sequence of directions using (non-matched) vicinity sampling in the upper rows and our proposed matched vicinity sampling in the lower rows.}
\end{figure}

In this section we will present some concrete examples for the differences in generator output dependent on the exact sample operation used to traverse the latent space of a generative model. To this end, the generator output for latent samples produced with linear interpolation, SLERP (spherical linear interpolation) of~\cite{white2016sampling} and our proposed matched interpolation will be compared. Please refer to Table~\ref{tab:operations} for an overview of the operators used in this Section.

\paragraph*{Setup }
We used DCGAN~\citep{DCGAN} generative models trained on LSUN bedrooms~\citep{yu2015lsun}, CelebA~\citep{liu2015faceattributes} and LLD~\citep{sagea2017icons}, an icon dataset, to qualitatively evaluate. For LSUN, the model was trained for two different output resolutions, providing $64\times64$ pixel and a $128\times128$ pixel output images (where the latter is used in figures containing larger sample images). The models for LSUN and the icon dataset where both trained on a uniform latent prior distribution, while for CelebA a Gaussian prior was used. The dimensionality of the latent space is 100 for both LSUN and CelebA, and 512 for the model trained on the icon model. Furthermore we use improved Wasserstein GAN (iWGAN) with gradient penalty~\citep{ImprovedWassersteinGAN} trained on CIFAR-10 at $32\times32$ pixels with a 128-dimensional Gaussian prior to produce the inception scores presented in Section~\ref{sec:quantitative}.

\paragraph*{2-point interpolation }
We begin with the classic example of 2-point interpolation: Figure~\ref{fig:2p-interp} shows three examples per dataset for an interpolation between 2 points in latent space. Each example is first done via linear interpolation, then SLERP and finally matched interpolation. In Figure~\ref{fig:2p-interp-detail} in the Appendix we show more densely sampled examples.

It is immediately obvious in Figures~\ref{fig:ip_icons} and \ref{fig:ip_lsun} that linear interpolation produces inferior results with generally more blurry, less saturated and less detailed output images. SLERP and matched interpolation are slightly different, however it is not visually obvious which one is superior. Differences between the various interpolation methods for CelebA (Figure~\ref{fig:ip_celebA}) are much more subtle to the point that they are virtually indistinguishable when viewed side-by-side. 
 This is not an inconsistency though: while distribution mismatch can cause large differences, it can also happen that the model generalizes well enough that it does not matter.

\paragraph*{Midpoint interpolation }
In all cases, the point where the interpolation methods diverge the most, is at the midpoint of the interpolation where $t=0.5$. Thus we provide 25 such interpolation midpoints in Figures~\ref{fig:icons-midpoints} (LLD icons) and~\ref{fig:lsun-midpoints} (LSUN) for direct comparison.

\paragraph*{4-point interpolation } 
An even stronger effect can be observed when we do 4-point interpolation, showcased in Figure~\ref{fig:lsun-4p} (LSUN) and Figure~\ref{fig:icons-4p} (LLD icons). The higher resolution of the LSUN output highlights the very apparent loss of detail and increasing prevalence of artifacts towards the midpoint in the linear version, compared to SLERP compared and our matched interpolation.

\paragraph*{Vicinity sampling }
Furthermore we provide two examples for vicinity sampling in Figures~\ref{fig:lsun_vinc} and \ref{fig:icons_vinc}. Analogous to the previous observations, the output under a linear operator lacks definition, sharpness and saturation when compared to both spherical and matched operators.

\paragraph*{Random walk }
An interesting property of our matched vicinity sampling is that we can obtain a \textit{random walk} in the latent space by applying it repeatedly: we start at a point $\y_0=\z$ drawn from the prior, and then obtain point $\y_i$ by sampling a single point in the vicinity of $\y_{i-1}$, using some fixed 'step size' $\epsilon$.

We show an example of such a walk in Figure~\ref{fig:random-walk}, using $\epsilon=0.5$.
As a result of the repeated application of the vicinity sampling operation, the divergence from the prior distribution in the non-matched case becomes stronger with each step, resulting in completely unrecognizable output images on the LSUN and LLD icon models.
Even for the CelebA model where differences where minimal before, they are quite apparent in this experiment. 
The random walk thus perfectly illustrates the need for respecting the prior distribution when performing any operation in latent space, as the adverse effects can cumulate through the repeated application of operators that do not comply to the prior distribution.

\subsection{Quantitative results}
\label{sec:quantitative}
We quantitatively confirm the observations of the previous section by using the Inception score\citep{salimans2016improved}. In Table~\ref{tab:inception_scores} we compare the Inception score of our trained models (i.e. using random samples from the prior) with the score when sampling midpoints from the 2-point and 4-point interpolations described above, reporting mean and standard deviation with 50,000 samples, as well as relative change to the original model scores if they are significant.
Compared to the original scores of the trained models, our matched operations are statistically indistinguishable (as expected) while  the linear interpolation gives a significantly lower score in all settings (up to 29\% lower). As observed for the quality visually, the SLERP heuristic gives similar scores to the matched operations.

\begin{table}
\vspace{-1.2cm}
\centering
\begin{tabular}{lllll}
\textbf{Dataset} & CIFAR-10 & LLD-icon & LSUN & CelebA\\ 
\textbf{Model} & iWGAN & DCGAN & DCGAN & DCGAN\\ 
\textbf{Prior} & Gaussian, 128-D & Uniform, 100-D & Uniform, 100-D &  Gaussian, 100-D\\ 
\textbf{Inception score} & $7.90 \pm 0.11$ & $3.70 \pm 0.09$ & $3.90 \pm 0.08$ & $2.05 \pm 0.04$\\ \hline \hline \\[-7pt]
\multicolumn{3}{c}{\textbf{Inception Score on midpoints of interpolation operations:}} \\ \hline  \\[-7pt]
2-point linear & $7.12 \pm 0.08$ \textcolor{gray}{(-10\%)} & $3.56 \pm 0.06$ \textcolor{gray}{(-4\%)} & $3.57 \pm 0.07$ \textcolor{gray}{(-8\%)} & $1.71 \pm 0.02$ \textcolor{gray}{(-17\%)}\\
2-point SLERP & $7.89 \pm 0.09$ & $3.68 \pm 0.09$ & $3.90 \pm 0.11$ & $2.04 \pm 0.04$\\
2-point matched & $7.89 \pm 0.08$ & $3.69 \pm 0.08$ & $3.89 \pm 0.08$ & $2.04 \pm 0.03$\\
4-point linear & $5.84 \pm 0.08$ \textcolor{gray}{(-26\%)} & $3.45 \pm 0.08$ \textcolor{gray}{(-7\%)} & $2.95 \pm 0.06$ \textcolor{gray}{(-24\%)} & $1.46 \pm 0.01$ \textcolor{gray}{(-29\%)}\\
4-point SLERP & $7.87 \pm 0.09$ & $3.72 \pm 0.09$ & $3.89 \pm 0.10$ & $2.04 \pm 0.04$\\
4-point matched & $7.91 \pm 0.09$ & $3.69 \pm 0.10$ & $3.91 \pm 0.10$ & $2.04 \pm 0.04$\\
\end{tabular}
\vspace{-0.2cm}
\caption{
        Inception scores on LLD-icon, LSUN, CIFAR-10 and CelebA for the midpoints of various interpolation operations. Scores are reported as mean $\pm$ standard deviation \textcolor{gray}{(relative change in \%)}.
}
\label{tab:inception_scores}
\end{table}

\section{Conclusions}
We have shown that the common latent space operations used for Generative Models induce distribution mismatch from the prior distribution the models were trained for.
This problem has been mostly ignored by the literature so far, partially due to the belief that this should not be a problem for uniform priors. 
However, our statistical and experimental analysis shows that the problem is real, with the operations used so far producing significantly lower quality samples compared to their inputs.
To address the distribution mismatch, we propose to use optimal transport to minimally modify (in $l_1$ distance) the operations such that they fully preserve the prior distribution.
We give analytical formulas of the resulting (matched) operations for various examples, which are easily implemented.
The matched operators give a significantly higher quality samples compared to the originals, having the potential to become standard tools for evaluating and exploring generative models.

\bibliography{iclr2018_conference}
\bibliographystyle{iclr2018_conference}

\newpage
\section{Appendix}
\subsection{On the curse of dimensionality and geometric outliers}
\label{app:curse}
We note that the analysis here can bee seen as a more rigorous version of an observation made by \cite{white2016sampling}, who experimentally show that there is a significant difference between the average norm of the midpoint of linear interpolation and the points of the prior, for uniform and Gaussian distributions.

Suppose our latent space has a prior with $\z=[Z_1,\cdots,Z_d]\in [-1,1]^d$ with i.i.d entries $Z_i\sim Z$. 
In this case, we can look at the squared norm 
\begin{align}
\|\z\|^2 = \sum_{i=1}^d Z_i^2.
\end{align}
From the Central Limit Theorem (CLT), we know that as $d\to \infty$,
\begin{align}
\sqrt{d}(\frac{1}{d}\|\z\|^2 - \mu_{Z^2} ) \rightarrow \mathcal{N}(0,\sigma_{Z^2}^2),
\end{align}
in distribution. Thus, assuming $d$ is large enough such that we are close to convergence, we can approximate the distribution of $\|\z\|^2$ as $\mathcal{N}(d \mu_{Z^2},d\sigma_{Z^2}^2)$.
In particular, this implies that almost all points lie on a relatively thin spherical shell, since the mean grows as $O(d)$ whereas the standard deviation grows only as $O(\sqrt{d})$.

We note that this property is well known for i.i.d Gaussian entries (see e.g. Ex. 6.14 in \cite{mackay2003information}). For Uniform distribution on the hypercube it is also well known that the mass is concentrated in the corner points (which is consistent with the claim here since the corner points lie on a sphere).

Now consider an operator such as the midpoint of linear interpolation, $\y=\frac{1}{2}\z_1+\frac{1}{2}\z_2$, with components $Y^{(i)}=\frac{1}{2}Z^{(i)}_1 + \frac{1}{2}Z^{(i)}_2$. Furthermore, let's assume the component distribution $p_Z$ is symmetric around $0$, such that $E[Z]=0$. 

In this case, we can compute:
\begin{align}
E[(Y^{(i)})^2]=\text{Var}[\frac{1}{2}Z^{(i)}_1+\frac{1}{2}Z^{(i)}_2]=\frac{1}{2}\text{Var}[Z]=\frac{1}{2}\mu_{Z^2}^2 \\
\text{Var}[(Y^{(i)})^2]=\text{Var}[(\frac{1}{2}Z^{(i)}_1+\frac{1}{2}Z^{(i)}_2)^2]=\frac{1}{4}\text{Var}[Z^2]=\frac{1}{4}\sigma_{Z^2}^2.
\end{align}
Thus, the distribution of $\|\y\|^2$ can be approximated with $\mathcal{N}(\frac{1}{2}d \mu_{Z^2},\frac{1}{4}d\sigma_{Z^2}^2)$.

Therefore, $\y$ also mostly lies on a spherical shell, but with a different radius than $\z$. In fact, the shells will intersect at regions which have a vanishing probability for large $d$. In other words, when looking at the squared norm $\|\y\|^2$, $\|\y\|^2$ is a (strong) outlier with respect to the distribution of  $\|\z\|^2$.

\subsection{Proof of Theorem~\ref{1dthm}}
\begin{proof}
We will show it for the Kantorovich problem, the Monge version is similar.

Starting from \eqref{eq:KP}, we compute
\begin{align}
&inf \left\{ E_{(\x,\y)\sim p_{\x,\y}}[c(\x,\y)] \Big|  (\x,\y)\sim p_{\x,\y}, \x\sim p_{\x}, \y\sim p_{\y} \right\}\\
&=inf \left\{ E_{(\x,\y)\sim p_{\x,\y}}[\sum_{i=1}^d  C(\x^{(i)},\y^{(i)})] \Big|  (\x,\y)\sim p_{\x,\y}, \x\sim p_{\x}, \y\sim p_{\y} \right\}\\
&=inf \left\{ \sum_{i=1}^d  E_{(\x,\y)\sim p_{\x,\y}}[C(\x^{(i)},\y^{(i)})] \Big|  (\x,\y)\sim p_{\x,\y}, \x\sim p_{\x}, \y\sim p_{\y} \right\}\label{branchpoint}\\
&\geq \sum_{i=1}^d  inf \left\{ E_{(\x,\y)\sim p_{\x,\y}}[C(\x^{(i)},\y^{(i)})] \Big|  (\x,\y)\sim p_{\x,\y}, \x\sim p_{\x}, \y\sim p_{\y} \right\}\label{indmin}\\
&= \sum_{i=1}^d  inf \left\{ E_{(X,Y)\sim p_{X,Y}}[C(X,Y)] \Big|  (X,Y)\sim p_{X,Y}, X\sim p_{X}, Y\sim p_{Y} \right\}\\
&= d \cdot inf \left\{ E_{(X,Y)\sim p_{X,Y}}[C(X,Y)] \Big|  (X,Y)\sim p_{X,Y}, X\sim p_{X}, Y\sim p_{Y} \right\},\\
\end{align}
where the inequality in \eqref{indmin} is due to each term being minimized separately.

Now let $\mathcal{P}_d(X,Y)$ be the set of joints $p_{\x,\y}$ with  $p_{\x,\y}(x,y)=\prod_{i=1}^dp_{X,Y}(x^{(i)},y^{(i)})$ where $p_{X,Y}$ has marginals $p_X$ and $p_Y$.
In this case $\mathcal{P}_d(X,Y)$ is a subset of all joints $p_{\x,\y}$ with marginals $p_{\x}$ and $p_{\y}$, where the pairs $(\x^{(1)},\y^{(1)}),\dots,(\x^{(d)},\y^{(d)}))$ are constrained to be i.i.d.
Starting again from \eqref{branchpoint} can compute:
\begin{align}
&inf \left\{ \sum_{i=1}^d  E_{(\x,\y)\sim p_{\x,\y}}[C(\x^{(i)},\y^{(i)})] \Big|  (\x,\y)\sim p_{\x,\y}, \x\sim p_{\x}, \y\sim p_{\y} \right\}\nonumber\\
&\leq inf \left\{ \sum_{i=1}^d  E_{(\x,\y)\sim p_{\x,\y}}[C(\x^{(i)},\y^{(i)})] \Big|  p_{\x,\y}\in\mathcal{P}_d(X,Y)  \right\}\label{smallerset}\\
&= inf \left\{ \sum_{i=1}^d  E_{(\x,\y)\sim p_{\x,\y}}[C(\x^{(i)},\y^{(i)})] \Big|  p_{\x,\y}\in\mathcal{P}_d(X,Y)  \right\}\\
&= inf \left\{ \sum_{i=1}^d  E_{(X,Y)\sim p_{X,Y}}[C(X,Y)] \Big| (X,Y)\sim p_{X,Y}, X\sim p_{X}, Y\sim p_{Y}  \right\}\\
&= d \cdot inf \left\{ E_{(X,Y)\sim p_{X,Y}}[C(X,Y)] \Big|  (X,Y)\sim p_{X,Y}, X\sim p_{X}, Y\sim p_{Y} \right\},\\
\end{align}
where the inequality in \eqref{smallerset} is due to minimizing over a smaller set.

Since the two inequalities above are in the opposite direction, equality must hold for all of the expressions above, in particular:
\begin{align}
&inf \left\{ E_{(\x,\y)\sim p_{\x,\y}}[c(\x,\y)] \Big|  (\x,\y)\sim p_{\x,\y}, \x\sim p_{\x}, \y\sim p_{\y} \right\}\\
&= d \cdot inf \left\{ E_{(X,Y)\sim p_{X,Y}}[C(X,Y)] \Big|  (X,Y)\sim p_{X,Y}, X\sim p_{X}, Y\sim p_{Y} \right\}
\end{align}
Thus, \eqref{eq:KP} and \eqref{eq:KP-1-D} equal up to a constant, and minimizing one will minimize the other. Therefore the minimization of the former can be done over $p_{X,Y}$ with $p_{\x,\y}(x,y)=\prod_{i=1}^dp_{X,Y}(x^{(i)},y^{(i)})$.
\end{proof}

\subsection{Calculations for Examples}
\label{app:examples}
In the next sections, we illustrate how to compute the matched operations for a few examples, in particular for linear interpolation and vicinity sampling, using a uniform or a Gaussian prior.
We picked the examples where we can analytically compute the uniform transport map, but note that it is also easy to compute $F^{[-1]}_Z$ and $(F_Y(y))$ numerically, since one only needs to estimate CDFs in one dimension.

Since the components of all random variables in these examples are i.i.d, for such a random vector $\x$ we will implicitly write $X$ for a scalar random variable that has the distribution of the components of $\x$.

When computing the monotone transport map  $T_{X\to Y}^{\text{mon}}$ , the following Lemma is helpful.
\begin{lemma}[Theorem 2.5 in \cite{santambrogio2015optimal}]\label{unique_monotone}
Suppose a mapping $g(x)$ is non-decreasing and maps a continuous distribution $p_X$ to a distribution $p_Y$, i.e.
        \begin{equation}
                g(X)\sim Y,
        \end{equation}
        then $g$ is the monotone transport map $T_{X\to Y}^{\text{mon}}$.
\end{lemma}
According to Lemma~\ref{unique_monotone}, an alternative way of computing $T_{X\to Y}^{\text{mon}}$ is to find some $g$ that is non-decreasing and transforms $p_X$ to $p_Y$. 

\subsubsection*{Example 1:Uniform Linear Interpolation}
\label{sec:matched_uniform}
Suppose $\z$ has uniform components $Z \sim \text{Uniform}(-1,1)$. In this case, $p_Z(z) = 1/2$ for $-1<z<1$.

Now let $\y_t=t\z_1 + (1-t)\z_2$ denote the linear interpolation between two points $\z_1,\z_2$, with component distribution $p_{Y_t}$.
Due to symmetry we can assume that $t>1/2$, since $p_{Y_t}=p_{Y_{1-t}}$.
We then obtain $p_{Y_t}$ as the convolution of $p_{tZ}$ and $p_{(1-t)Z}$, i.e. $p_{Y_t}=p_{tZ}*p_{(1-t)Z}$.
First we note that $p_{tZ}=1/(2t)$ for $-t<z<t$ and  $p_{(1-t)Z}=1/(2(1-t))$ for $-(1-t)<z<1-t$.
We can then compute:
\begin{align}
&p_{Y_t} (y) = (p_{tZ}*p_{(1-t)Z})(y)\\
&=\frac{1}{2(1-t)(2t)}\begin{cases}
0    &\text{ if } y < -1 \\
y+1  &\text{ if } -1 < y < -t + (1-t) \\
2-2t &\text{ if } -t + (1-t) < y < t - (1-t) \\
-y+1 &\text{ if } t - (1-t) < y < 1 \\
0    &\text{ if }1 < y  \\
\end{cases} \\
\end{align}

The CDF $F_{Y_t}$ is then obtained by computing
\begin{align}
        &F_{Y_t}(y) = \int_{-\infty}^y p_{Y_t}(y') dy' \\
 &= \frac{1}{2(1-t)(2t)}\begin{cases}
0    &\text{ if } y < -1 \\
\frac{1}{2}(y+1)(y+1)  &\text{ if } -1 < y < 1-2t \\
2(1-t)(y+t)&\text{ if } 1-2t < y < 2t-1 \\
2(1-t)(3t-1) +(-\frac{1}{2}y^2+y + \frac{1}{2}(2t-1)^2-(2t-1))  &\text{ if } 2t-1 < y < 1 \\
2(1-t)(2t)   &\text{ if }1 < y  \\
\end{cases}
\end{align}

Since $p_Z(z)=1/2$ for $|z|<1$, we have $F_Z(z)=\frac{1}{2}z+\frac{1}{2}$ for $|z|<1$. This gives $F^{[-1]}_Z(p)=2(p-\frac{1}{2})$.

Now, we just compose the two mappings to obtain $T_{Y_t\to Z}^{\text{mon}}(y)=F^{[-1]}_Z(F_{Y_t}(y))$.

\subsubsection*{Example 2: Uniform Vicinity Sampling and Random Walk}
\label{sec:random-walk}
Let $\z$ again have uniform components on $[-1,1]$. For vicinity sampling, we want to obtain new points $\z'_1,\cdot,\z'_k$ which are close to $\z$.
We thus define 
\begin{equation}
        \z'_i := \z + \epsilon \u_i,
\end{equation} where $\u_i$ also has uniform components, such that each coordinate of $\z'_i$ differs at most by $\epsilon$ from $\z$.
By identifying $t Z'_i  = tZ + (1-t)U_i$ with $t=1/(1+\epsilon)$, we see that $tZ'_i$ has identical distribution to the linear interpolation $Y_t$ in the previous example.
Thus $g_t(Z'_i):=T_{Y_t\to Z}^{\text{mon}}(tZ'_i)$ will have the distribution of $Z$, and by Lemma\ref{unique_monotone} is then the monotone transport map from $Z'_i$ to $Z$.

\subsubsection*{Example 3: Gaussian Linear Interpolation, Vicinity Sampling and Analogies}
\label{sec:matched_gauss}
Suppose $\z$ has components $Z\sim \mathcal{N}(0,\sigma^2)$.
In this case, we can compute linear interpolation as before, $\y_t = t\z_1 + (1-t)\z_2$.
Since the sum of Gaussians is Gaussian, we get, $Y_t \sim \mathcal{N}(0,t^2\sigma^2 + (1-t)^2 \sigma^2)$. 
Now, it is easy to see that with a proper scaling factor, we can adjust the variance of $Y_t$ back to $\sigma^2$.
That is, $\frac{1}{\sqrt{t^2 + (1-t)^2}} Y_t \sim \mathcal{N}(0,\sigma^2)$, so by Lemma~\ref{unique_monotone} $g_t(y) := \frac{1}{\sqrt{t^2 + (1-t)^2}}y$ is the monotone transport map from $Y_t$ to $Z$.

By adjusting the vicinity sampling operation to 
\begin{equation}
        \z'_i := \z + \epsilon \e_i,
\end{equation} where $\e_i\sim\mathcal{N}(0,1)$, we can similarly find the monotone transport map $g_{\epsilon}(y)=\frac{1}{\sqrt{1+\epsilon^2}} y$.

Another operation which has been used in the literature is the ``analogy'', where from samples $\z_1,\z_2,\z_3$, one wants to apply the difference between $\z_1$ and $\z_2$, to $\z_3$. The transport map is then $g(y)=\frac{1}{\sqrt{3}} y$

\newpage
\subsection{Additional experiments}

\begin{figure}[htp]
	\centering
	\begin{subfigure}[t]{0.33\linewidth}
		\centering\includegraphics[width=0.95\linewidth]{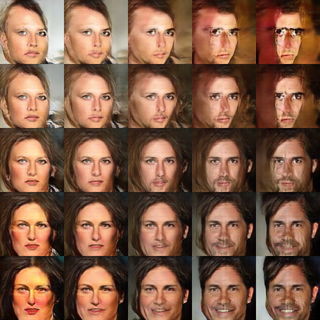}
		\caption{\label{fig:celebA-4p-linear} Linear interpolation}
	\end{subfigure}%
	\begin{subfigure}[t]{0.33\linewidth}
		\centering\includegraphics[width=0.95\linewidth]{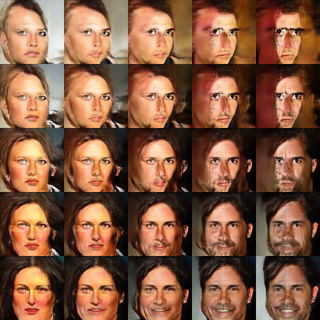}
		\caption{\label{fig:celebA-4p-slerp} Spherical linear interpolation}
	\end{subfigure}
	\begin{subfigure}[t]{0.33\linewidth}
		\centering\includegraphics[width=0.95\linewidth]{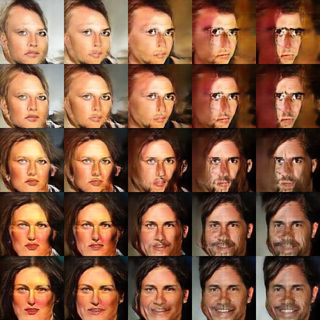}
		\caption{\label{fig:celebA-4p-matched} Distribution matched}
	\end{subfigure}
	\caption{\label{fig:celebA-4p} 4-point interpolation between 4 sampled points (corners) from DCGAN trained on CelebA with Gaussian prior. The same interpolation is shown using linear, SLERP and distribution matched interpolation.}
\end{figure}

\begin{figure}[htp]
	\centering
	\begin{subfigure}[t]{0.33\linewidth}
		\centering\includegraphics[width=0.95\linewidth]{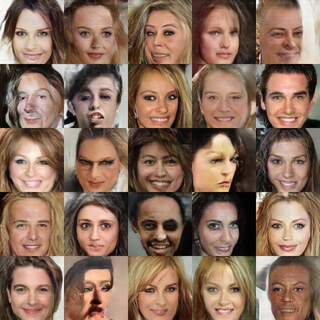}
		\caption{\label{fig:celebA-midpoints-linear} Linear}
	\end{subfigure}%
	\begin{subfigure}[t]{0.33\linewidth}
		\centering\includegraphics[width=0.95\linewidth]{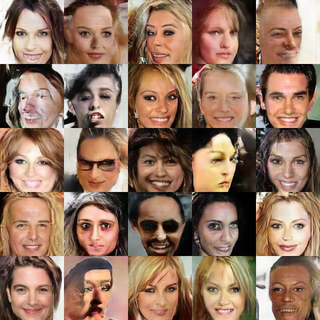}
		\caption{\label{fig:celebA-midpoints-slerp} Spherical}
	\end{subfigure}
	\begin{subfigure}[t]{0.33\linewidth}
		\centering\includegraphics[width=0.95\linewidth]{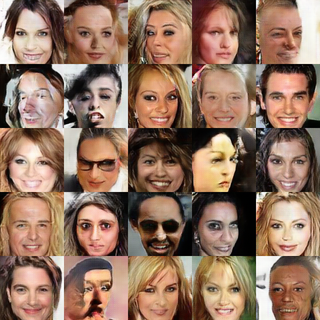}
		\caption{\label{fig:celebA-midpoints-matched} Distribution matched}
	\end{subfigure}
	\caption{\label{fig:celebA-midpoints}Midpoint sampling for linear, SLERP and uniform-matched interpolation when using the same pairs of sample points on CelebA with Gaussian prior.}
\end{figure}

\begin{figure}[htp]
	\centering
	\begin{subfigure}[t]{0.33\linewidth}
		\centering\includegraphics[width=0.8\linewidth]{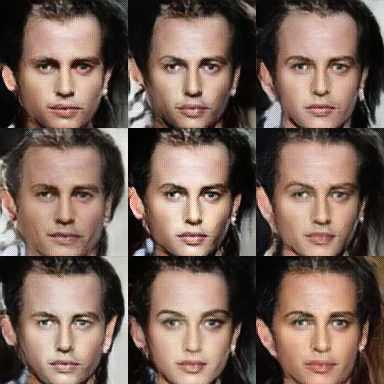}
		\caption{\label{fig:celebA_vinc_linear} Vicinity sampling}
	\end{subfigure}%
	\begin{subfigure}[t]{0.33\linewidth}
		\centering\includegraphics[width=0.8\linewidth]{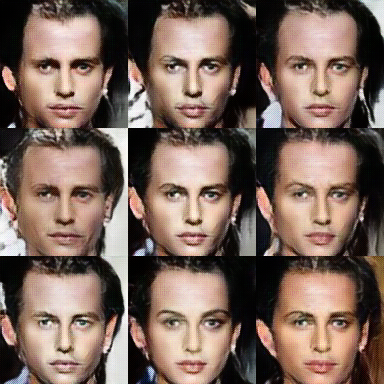}
		\caption{\label{fig:celebA_vinc_matched} Matched vicinity sampling}
	\end{subfigure}
	\caption{\label{fig:celebA_vinc} Vicinity sampling on CelebA dataset with Gaussian prior. The sample in the middle is perturbed in random directions producing the surrounding sample points.}
\end{figure}

\begin{figure}[tb]
	\centering
	\begin{subfigure}[b]{0.99\linewidth}
		\centering
		\includegraphics[width=0.95\linewidth]{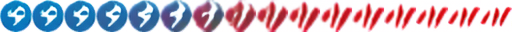}\\
		\includegraphics[width=0.95\linewidth]{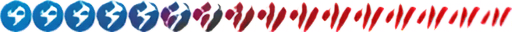}\\
		\includegraphics[width=0.95\linewidth]{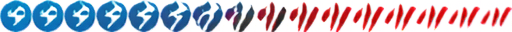}\\
		\includegraphics[width=0.95\linewidth]{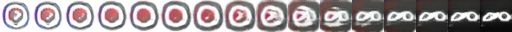}\\
		\includegraphics[width=0.95\linewidth]{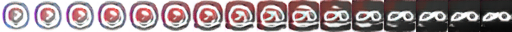}\\
		\includegraphics[width=0.95\linewidth]{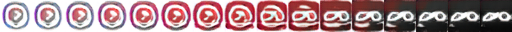}\\
		\includegraphics[width=0.95\linewidth]{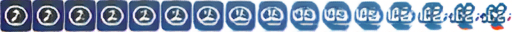}\\
		\includegraphics[width=0.95\linewidth]{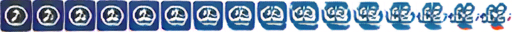}\\
		\includegraphics[width=0.95\linewidth]{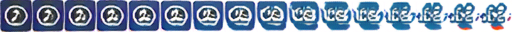}
		\caption{\label{fig:ip_icons_detail} LLD icon dataset}
	\end{subfigure}%

	\begin{subfigure}[b]{0.99\linewidth}
		\centering
		\includegraphics[width=0.95\linewidth]{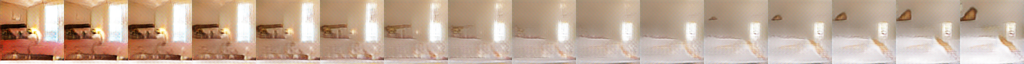}\\
		\includegraphics[width=0.95\linewidth]{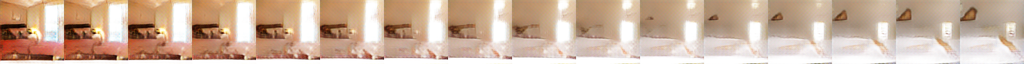}\\
		\includegraphics[width=0.95\linewidth]{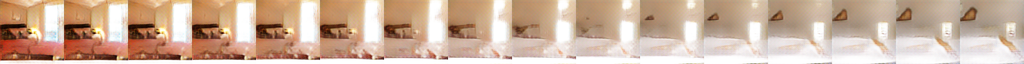}\\
		\includegraphics[width=0.95\linewidth]{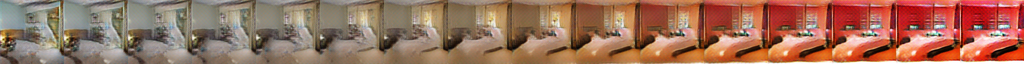}\\
		\includegraphics[width=0.95\linewidth]{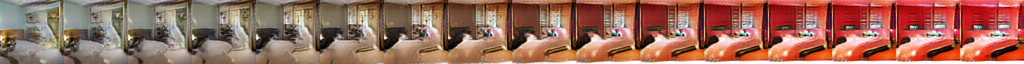}\\
		\includegraphics[width=0.95\linewidth]{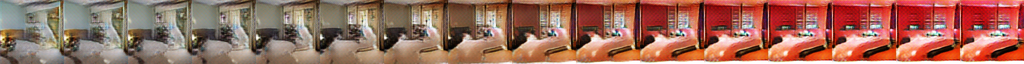}\\
		\includegraphics[width=0.95\linewidth]{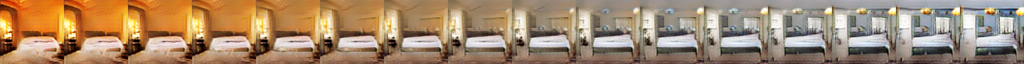}\\
		\includegraphics[width=0.95\linewidth]{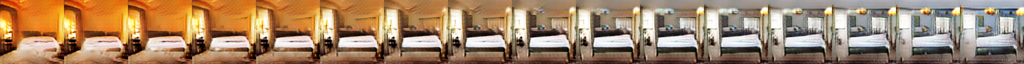}\\
		\includegraphics[width=0.95\linewidth]{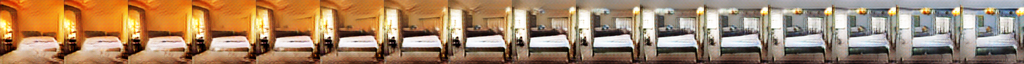}
		\caption{\label{fig:ip_lsun_detail} LSUN dataset}
	\end{subfigure}
	\caption{\label{fig:2p-interp-detail}2-point interpolation in detail: Each example shows linear, SLERP and transport matched interpolation from top to bottom respectively, with 16 points taken from the path, produced with DCGAN using a uniform prior distribution.}
\end{figure}

\end{document}